\documentclass[conference]{IEEEtran}
\usepackage{array}
\usepackage{booktabs}
\usepackage{multirow}
\usepackage{bm}
\usepackage{cite}
\usepackage{amsmath,amssymb,amsfonts}
\usepackage{algorithmic}
\usepackage{graphicx}
\usepackage{subfigure}
\usepackage{textcomp}
\usepackage{xcolor}
\usepackage[normalem]{ulem}
\usepackage{hyperref}
\usepackage{url}
\def\BibTeX{{\rm B\kern-.05em{\sc i\kern-.025em b}\kern-.08em
    T\kern-.1667em\lower.7ex\hbox{E}\kern-.125emX}}
\usepackage{soul}
\usepackage{algorithm}
\usepackage{algorithmic}

\usepackage{bm}

\usepackage{authblk}

\usepackage{ifthen}
\newboolean{authnotes}
\ifthenelse{\boolean{authnotes}}
{
\newcommand{\cl}[1]{\footnote{\color{orange}{\bf CL: #1}}}
\newcommand{\jz}[1]{\footnote{\color{purple}{\bf JZ: #1}}}
\newcommand{\lw}[1]{\footnote{\color{blue}{\bf LW: #1}}}

\usepackage{draftwatermark}
\SetWatermarkLightness{0.95}
\SetWatermarkScale{1}
}
{
\newcommand{\cl}[1]{}
\newcommand{\jz}[1]{}
\newcommand{\lw}[1]{}

}

\begin{document}

\title{Temporal Convolution Domain Adaptation Learning for Crops Growth Prediction}

\author[1]{Shengzhe Wang}
\author[1]{Ling Wang}
\author[1]{Zhihao Lin}
\author[2]{Xi Zheng}

\affil[1]{Department of Computer Science and Technology, Harbin Insititute of Technology }
\affil[2]{Department of Computing, Macquaire University}
\affil[1]{\textit{1180301010@stu.hit.edu.cn,wangling@hit.edu.cn,21S10325@stu.hit.edu.cn}}
\affil[2]{\textit{james.zheng@mq.edu.au}}


\maketitle
\pagestyle{plain}

\begin{abstract}
Existing Deep Neural Nets on crops growth prediction mostly rely on availability of a large amount of data. In practice, it is difficult to collect enough high-quality data to utilize the full potential of these deep learning models. In this paper, we construct an innovative network architecture based on domain adaptation learning to predict crops growth curves with limited available crop data. This network architecture overcomes the challenge of data availability by incorporating generated data from the developed crops simulation model. We are the first to use the temporal convolution filters as the backbone to construct a domain adaptation network architecture which is suitable for deep learning regression models with very limited training 
data of the target domain. We conduct experiments to test the performance of the network and compare our proposed architecture with other state-of-the-art methods, including a recent LSTM-based domain adaptation network architecture. The results show that the proposed temporal convolution-based network architecture outperforms all benchmarks not only in accuracy but also in model size and convergence rate.

\end{abstract}

\begin{IEEEkeywords}Pervasive Computing Applications, Deep Learning, Transfer Learning 

\end{IEEEkeywords}

\section{Introduction}\label{sec:intro}

Environments have enormous influences on crop growth and development, resulting in significant variation in crop yields. Under such conditions, the significance of crop growth prediction is threefold. Firstly, once the growth curve of crops can be correctly predicted, agricultural experts can make better planting decisions and thus increase agricultural output. Secondly, an accurate growth prediction gives farmers the advantage of beneficial marketing plans for their products. Thirdly, growth prediction before harvest is crucial for national food security \cite{horie1992yield} including policies concerning import/export plans and prices.
Therefore, crop growth prediction during crop production processes is essential for optimum crop management \cite{li2017oryza2000}.

The crop growth simulation model like WOFOST \cite{van1989wofost}, DSSAT \cite{jones2003dssat} have been widely used for this purpose. These models are based on the inherent laws of crop growth and development. Consequently, it can quantitatively describe and predict crop growth and the dynamic relationship between the environment and crops. The accuracy of these models is limited by assumptions that simplify the growth process and ignore some determining factors that are difficult to compute. With the development of deep neural networks, deep learning models can predict the growth curves of crops \cite{WHISLER1986141}\cite{wang2020deep}\cite{10.1145/3209811.3212707}\cite{VANKLOMPENBURG2020105709}, making them play an essential role in predicting agricultural production \cite{van1989wofost}\cite{jones2003dssat}\cite{keating2003overview} and smart agricultural decision-making \cite{cambra2019smart}\cite{salam2019internet}.

However, deep learning models are data-driven models which require a large amount of data to produce a precise prediction. State-of-the-art deep learning architectures \cite{sun2019county}\cite{khaki2020cnn} usually use accumulated data in specific regions, expanding to dozens of years. Some parts of these datasets are not even accessible online \cite{wang2020winter}.

In real-world applications, environment and crop conditions, including climate, soil, and seed, usually exhibit different data distribution among different regions. Many regions lack years of accumulated crops data. Adapting prediction models to a real farm on a smaller scale requires a large amount of historical crop data. Nevertheless, the challenge is that field experiments, and crop data collection are expensive and time-consuming as the growth of most crops expands for several months.

We address the challenge of data availability by combining simulation and data-driven models with innovative transfer learning architecture. To obtain the model that can effectively predict the growth curve of actual crops, we first use the simulation model to generate a large amount of crop growth data as our source domain dataset. Those simulated data reflect generic patterns of crop growth. However, we hypothesize that even though these data have a different distribution from real-world crop data, the knowledge learned from such data can help improve the performance of data-driven models, like deep neural networks. Correspondingly, we collect data on a real-world farm, and a small amount of actual crop growth data is obtained as our target domain dataset. We use the data in these two domains to carry out domain adversarial transfer learning and train a crop growth prediction model. 

Considering the gap between distributions of the source domain and target domain, we design a Temporal Convolutional Networks (TCN) backboned Domain Adversarial Neural Networks (DANN) architecture that can predict plant growth curves accurately with minimal training data of the target domain. Adversarial training of different parts in our architecture results in domain-invariant performance between the source domain and target domain datasets. Additionally, we select TCN as the feature extractor of the proposed architecture. TCN employs casual and dilated convolutions, increasing the reception field within fewer convolution layers. Concretely, the contributions of the paper are listed as follows:

\begin{itemize}
     
     \item A customized version of Domain Adversarial Neural Network in which a gradient reversal layer is used to generate the domain-invariant features is proposed for crop growth prediction in a data-limited situation, and a Gaussian distribution regressor is used to improve the model performance.
     \item Temporal Convolution Networks is used as the backbone of our network architecture to fully explore both temporal space (dilation) and spatial space (residual).
     \item Real-world experiments are conducted to evaluate our proposed network architecture, in which we collect target domain data from edge devices with multiple sensors and use a simulation platform to generate source data.
     
\end{itemize}
In our evaluation, we show that compared to other state-of-the-art benchmarks, our TCN-based network architecture performs much better in terms of prediction loss, model size and convergency rate.  

The structure of this paper is as follows. Section \ref{sec:relatedwork} presents background and related work. In Section \ref{sec:data}, we will describe the collection and process of our dataset. Section \ref{sec:solution} gives out specific problem and corresponding solutions. In Section \ref{sec:experiment}, we will give the specific results and analysis of the experiment to validate our proposals. Finally we will summarize our work in Section \ref{sec:conclusion}.

\section{Related work}\label{sec:relatedwork}

\subsection{Crop Growth Prediction}
Machine learning (ML) and Deep Learning (DL) have been used in crop growth prediction.
Representative works for ML method include artificial Neural Networks (Drummond \emph{et al.} \cite{drummond2003statistical}; Fortin \emph{et al.}\cite{fortin2011site}), support vector regression (Ruß\cite{russ2009data}), and k-nearest neighbor (Zhang et al. \cite{zhang2010simulation}).
There is also a comparative study of ANN, SVR, M5-Prime, kNN ML techniques, and Multiple Linear Regression for crop yield prediction in ten crop datasets\cite{gonzalez2014predictive}.

More recently, DL techniques have been applied for crop yield prediction.
There are some works such as multi-layer perceptron (MLP) \cite{niedbala2019application} and
LSTM models  \cite{schwalbert2020satellite}\cite{jiang2020deep}\cite{zhang2020combining}.
The idea of combining two kinds of network architecture also became popular recently.
For instance, CNN-LSTM based models were used in soybean yield prediction \cite{sun2019county} and winter wheat yield prediction \cite{wang2020winter}.
In another work \cite{khaki2020cnn}, a CNN-RNN model has been used for yield prediction based on environmental data and management practices.

However, all ML and DL based methods require a considerable amount of data. Due to the characteristics of agricultural production, collecting a large amount of data is both time-consuming and expensive. Furthermore, labeled training data required by these methods is quite scarce. Firstly, agricultural production is a seasonal work that leads to long-term data collection. Furthermore, a large amount of data collected by sensors in the farm are not labeled. What is more, open-source data in the agriculture field are much fewer than in other scenes, such as auto-driving. Due to the harsh environment of the farm, continuous data collection using sensors is rather challenging. Accidents like thunder or power failure, even agricultural vehicles running into a sensor node can easily damage the data collection process.

\subsection{Domain Adaptation}

Domain Adaptation (DA) is a particular case of transfer learning that leverages labeled data in one or more related source domains to learn a classifier for unseen or unlabeled data in a target domain. Many shallow domain adaption methods have been studied in the transfer learning field, including instance re-weighting \cite{zadrozny2004learning}\cite{kanamori2008efficient}, parameter adaptation \cite{yang2007cross}\cite{duan2012exploiting}, feature augmentation \cite{daume2009frustratingly}, feature space alignment \cite{fernando2013unsupervised}, and unsupervised feature transformation \cite{pan2010domain}.

In deep domain adaptation, fine-tuning is widely used. Yosinskiet \emph{et al.} \cite{yosinski2014transferable} conducted a study on the impact of transferability of features from different layers of the network. It is demonstrated that lower layer features are typically more general (i.e., class agnostic in the context of classification) while higher-layer features have greater specificity than lower ones.
Another promising deep domain adaptation method is using an adversarial discriminative model.
The two most widely used models are the Domain-Adversarial Neural Networks (DANN) \cite{ganin2016domain} and the Adversarial Discriminative Domain Adaptation (ADAA)~\cite{tzeng2017adversarial}.
DANN~\cite{ganin2016domain} integrates a domain classifier into the standard architecture to promote the emergence of domain-independent features that are discriminative for the main learning task on the source domain and indiscriminate concerning the data distribution shift between the domains.
ADAA \cite{tzeng2017adversarial} uses an inverted label GAN loss to split the optimization into two independent objectives for the generator and the discriminator. This model considers the independent source and target mappings, allowing domain-specific feature extraction. Also, the target weights are initialized by the network pretrained on the source.
Compared to adversarial discriminative models, generative adversarial networks combine the discriminative model with a generative component in general based on GANs \cite{goodfellow2014generative}.

In this paper, due to the lack of historical data on the target domain, we chose to approach the problem with domain adaptation, allowing the transfer of knowledge from the source domain to the target domain. We specifically chose DANN for its appealing performance in domain adaptation transfer learning. Additionally, DANN integrates domain adversarial training of neural networks in a single process, simplifying the implementation and lowering the parameters search cost.

\subsection{Temporal Convolutional Networks for Sequence Modeling}

Colin Lea \emph{et al.} \cite{lea2017temporal} first introduces Temporal Convolutional Networks (TCN), which use dilation convolution and residual connect to expand the receptive field exponential, thus capturing long-range temporal patterns with fewer layers. Their work achieves state-of-the-art performance on challenging datasets and outperformed strong baselines, including Bidirectional LSTM.
Then Shaojie Bai \emph{et al.} \cite{bai2018empirical} presented an empirical evaluation of generic convolutional and recurrent architectures across a comprehensive suite of sequence modeling tasks. Their results indicate that TCN models outperform generic recurrent architectures such as LSTMs and GRUs.

It has shown that temporal convolutional architectures have been successfully used in some scenarios, such as audio synthesis and machine translation. 
In our work, to the best of our knowledge, we are the first to leverage TCN with DANN for crop prediction and analyze the potential of such a combination for pervasive systems with similar constraints (deep learning models training with limited datasets).

\section{Data Collection and Pre-processing} \label{sec:data}
\subsection{Data Collection}
The farm where we collect real crops data is located in the northeast of China. This farm covers about 250 acres. The main crops grown on this farm are corn and rice. Four sets of edge nodes are deployed on the farm for field environmental monitoring, each containing 14 different kinds of sensors to monitor different environmental variables. These sensors include soil temperature and humidity sensor, air temperature and humidity sensor, light intensity sensor, wind sensor, rain sensor, and air pressure sensor.
The features and unit of measure for these sensors are shown in Table \ref{table:envi}. The environmental monitoring process has started in September 2020 and has continued to the present. Each edge node uploads the environmental variables every half hour to the data center, and we now have over 70000 records in total. Each record contains 16 columns, including 14 features, collection time, and edge device number.
For crop images, three spherical cameras are deployed on the farm. A program is developed to drive the spherical camera and deployed in a Raspberry Pi 3B model. Each camera is set to take photos from several preset points every hour. They have taken over 40000 pictures, including corn and rice, from May 2021 to October 2021.

Sensor data is directly sent to a Mysql server deployed on the cloud for data transmission and storage. Crop pictures are sent to an Object Storage Service (OSS) hosted in the cloud for generating their respective URLs. Then the URLs combined with the image capture time and location information are also inserted into the Mysql server mentioned above. All network transmissions are done using a 4G network.

The actual data collection process is far more complicated than anticipated. Several power failure accidents appear due to the changeable environment on the farm. The Power mode for the edge nodes has to be changed from solar to AC. The sandy and dusty environment have also caused some trouble in collecting data. These issues also strengthen our assumption that high-quality data is challenging to retrieve, and after the data cleaning process, less data remains for model training.

\begin{figure}
    \centering
    \includegraphics[width=0.65\linewidth]{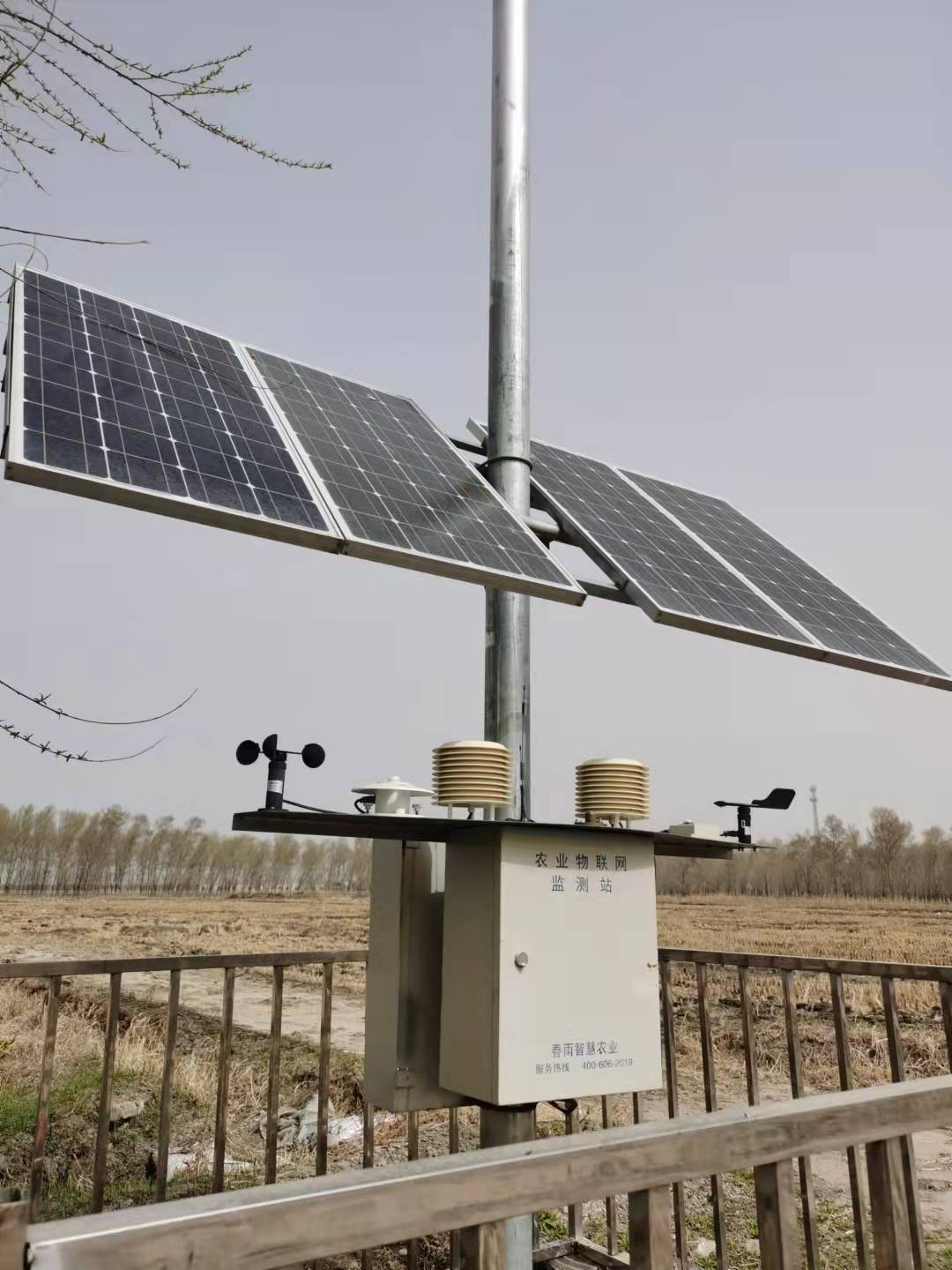}
    \caption{Sensor node located on the real farm}
    \label{fig:device}
\end{figure}

\begin{figure*}[htbp]
\centering
\subfigure[corn image taken in 2021-06-29]{
\begin{minipage}[t]{0.48\linewidth}
\centering
\includegraphics[width=0.98\textwidth]{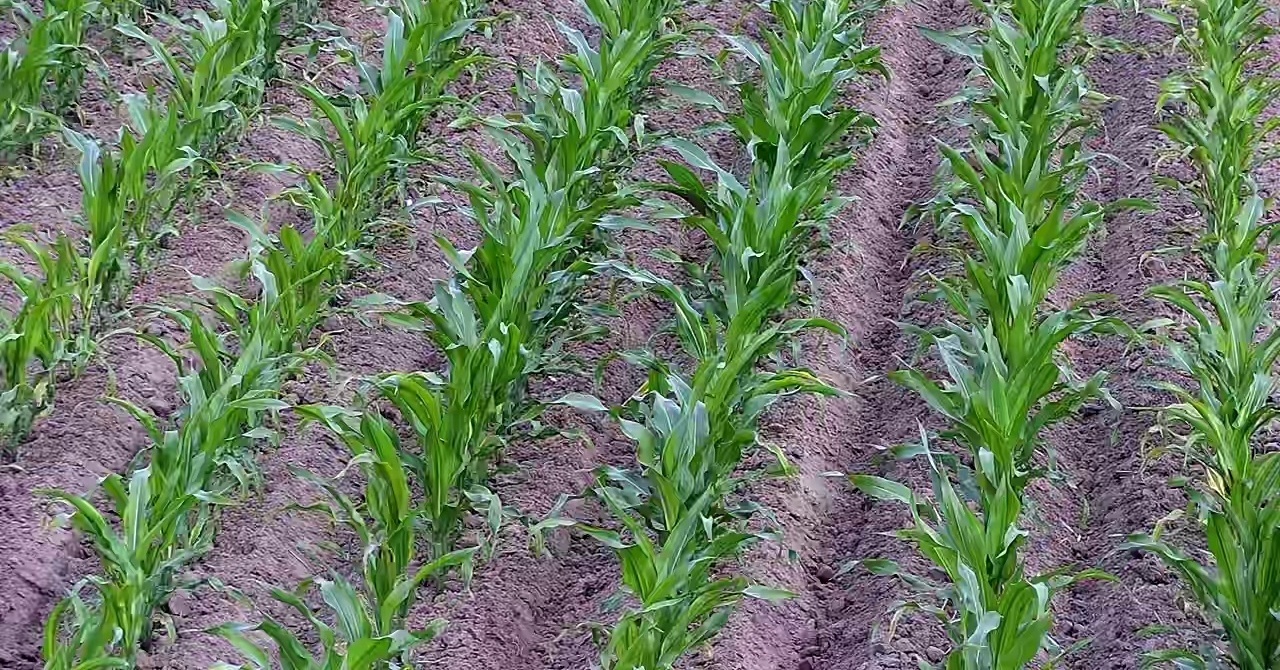}
\end{minipage}%
}
\subfigure[corn image taken in 2021-07-06]{
\begin{minipage}[t]{0.48\linewidth}
\centering
\includegraphics[width=0.98\textwidth]{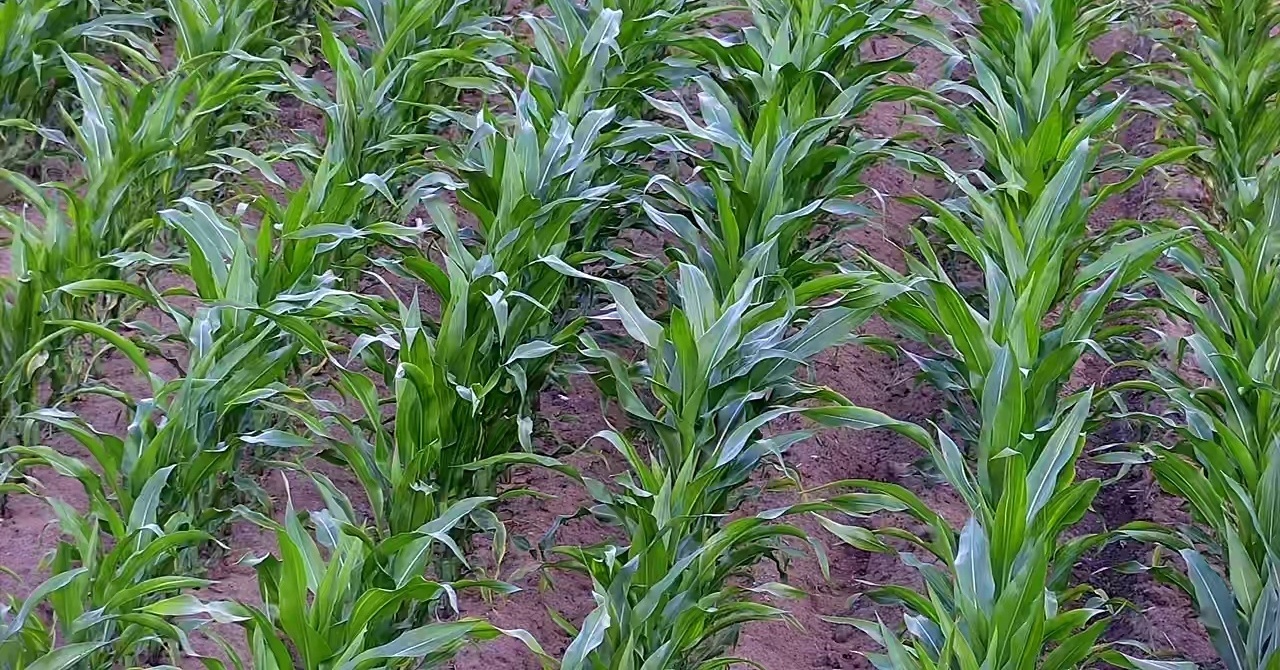}
\end{minipage}%
}
\centering
\caption{Example of image data automatically collected by camera at the same preset point}
\end{figure*}
\begin{table}[]
\centering
\caption{Environmental variables being monitored}

\label{table:envi}
\setlength{\tabcolsep}{7mm}
\renewcommand\arraystretch{1.1}
\begin{tabular}{l|l}
\hline
Features          & measurement          \\ \hline
Air Temperature   & degree centigrade/°C \\
Humidity          & percentage           \\
Illuminance       & lux                  \\
Soil Temperature  & degree centigrade/°C \\
soil humidity     & percentage           \\
Air Pressure      & HPa                  \\
Rainfall          & mm                   \\
Wind Speed        & m/s                  \\
Wind Direction    & E, S, W, N           \\
Photosynthetic    & $\mu$mol*$m^-2$*$s^-1$\\
Sun Exposure Time & hour                 \\
Carbon Dioxide    & ppm                  \\
Soil Salinity     & ms/cm                \\
Soil PH           & 1                    \\ \hline
\end{tabular}
\centering
\end{table}

\subsection{Data Pre-processing}
For the source domain, we develop a simulation model to generate a large amount of crop data. Specifically, Python Crop Simulation Environment (PCSE), a python implementation of WOFOST \cite{van1989wofost}, is used. A default file containing parameters for maize corn is selected, and open source weather data \cite{AnnualClimateChina19812010} is used to generate simulated crop data as our source domain dataset. Specifically, the PCSE simulates a maize growth cycle. The virtual maize crop is set to be sowed on May 1st and harvested on October 20th. As a result, the total time of simulated growth is 173 days. We generate over 400 rounds of the complete growth process of virtual maize corn. The output of the PCSE is considered as time-series data where each element in the series is a multi-dimensional vector. The vector represents all the features concerning weather conditions and the crop itself, including Leaf Area Index (LAI) which characterizes the plant growth situation. It is defined as the green leaf area per unit ground surface area, and it serves as an essential feature for crop growth prediction.

For the target domain, as mentioned above, data are collected from a real-world farm. Fourteen kinds of sensors periodically record data that comprise our target domain dataset. Images of the growing crop will be converted to the LAI to manifest the growing situation of the crop. 

Image data is manually labeled for accuracy to compute LAI. Since electrical shortage frequently happens on the farm, a specific range of our collected data is selected as testing data from June 29th to July 9th in 2021. The crop images have better quality in this period without electric interrupt. 

Several common features are selected in both domains for the same feature space, and the merged time series dataset is constructed at a regular interval of 24 hours. We also cut the longer simulated data into a shorter range to construct the same feature space. Specifically, the highest temperature, the lowest temperature, average humidity, and the average irradiation in a day are used as features of weather data at an interval of a day. LAI is used as the label and input at the same time. The LAI value serving as the label in a previous day will be the input feature of the crop growth situation for the next day. 

In general, after merging two domains, each batch of data has a length of 11 days. Each day has five features as LAI, the highest temperature, the lowest temperature, the average humidity, and the average irradiation power together with one label which is the LAI of the next day. There are over 6000 batches in the source domain dataset and mere 46 batches in the target domain dataset.

\section{Solution} \label{sec:solution}

\begin{figure}[htbp]
\centerline{\includegraphics[width=0.40\textwidth]{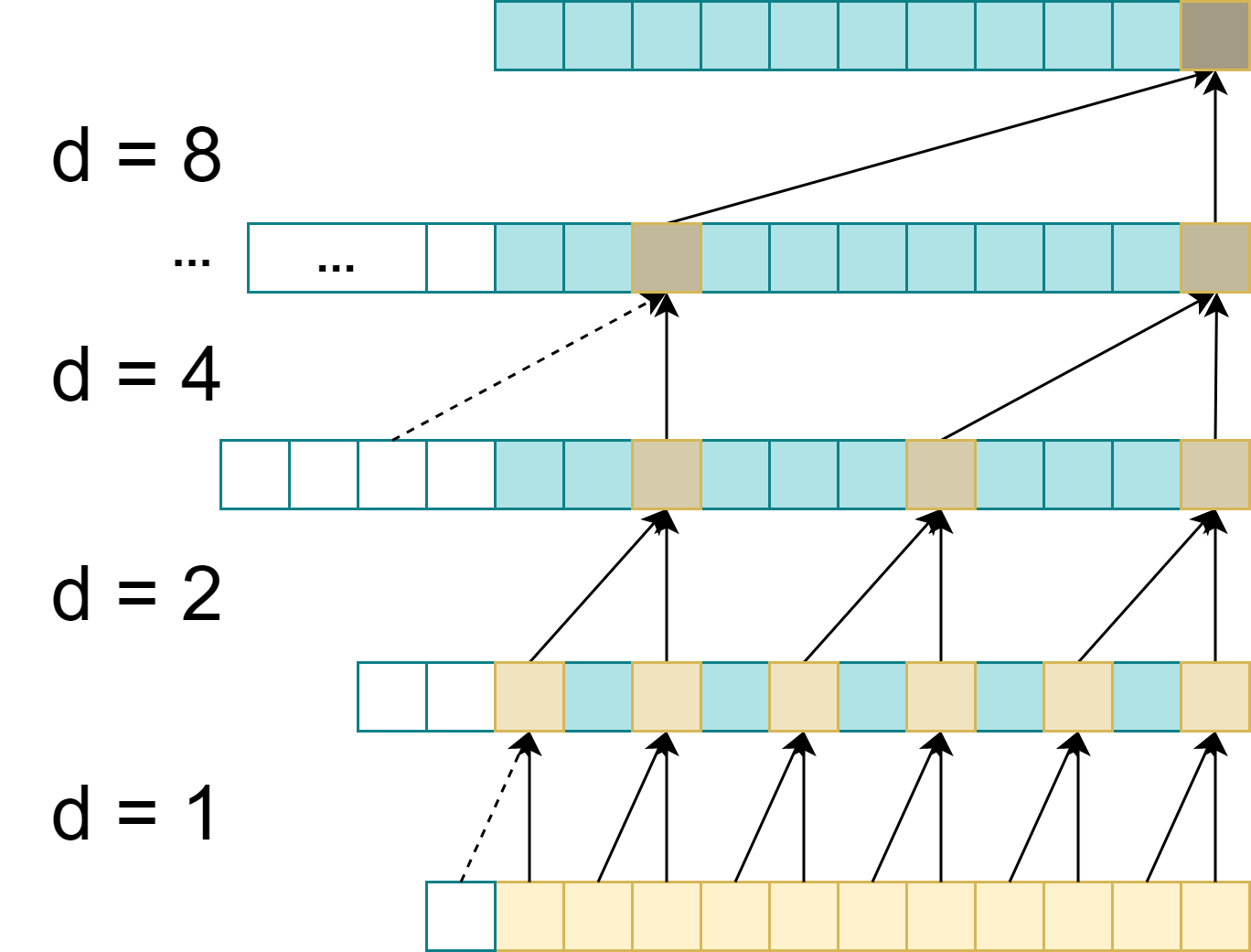}}
\caption{Dilated Convolution in TCN }
\label{fig:Dilated Convolution}
\end{figure}

Traditional transfer learning mainly relies heavily on fine-tuning, which re-trains parameters on the target domain for a pre-trained net. However, fine-tuning is still far from satisfied in the presence of a distribution shift between source and target domain. Ganin \emph{et al.} \cite{ganin2016domain} proposed an architecture that emphasizes features that cannot be discriminated between domains. They constructed an additional domain classifier to train the feature extractor, usually a convolution network, to generate domain-invariant features. In this manner, the distribution of features learned from the source domain will play a more critical role in transferring the network to the target domain. This domain adaptation architecture, called Domain-Adversarial Neural Networks (DANN), employs a gradient reversal layer to simplify the implementation of the reversed gradient updates within the existing deep learning framework. In this case, an adversary between these two parts results in a feature extractor that tends to provide better features without domain characteristics. The previous study \cite{ganin2016domain}\cite{shao2021few-shot} shows that DANN can train a feature extractor with the least domain characteristics, thus providing better performance in the target domain.

\begin{figure*}[htbp]
\leftline{\includegraphics[width=0.96\textwidth]{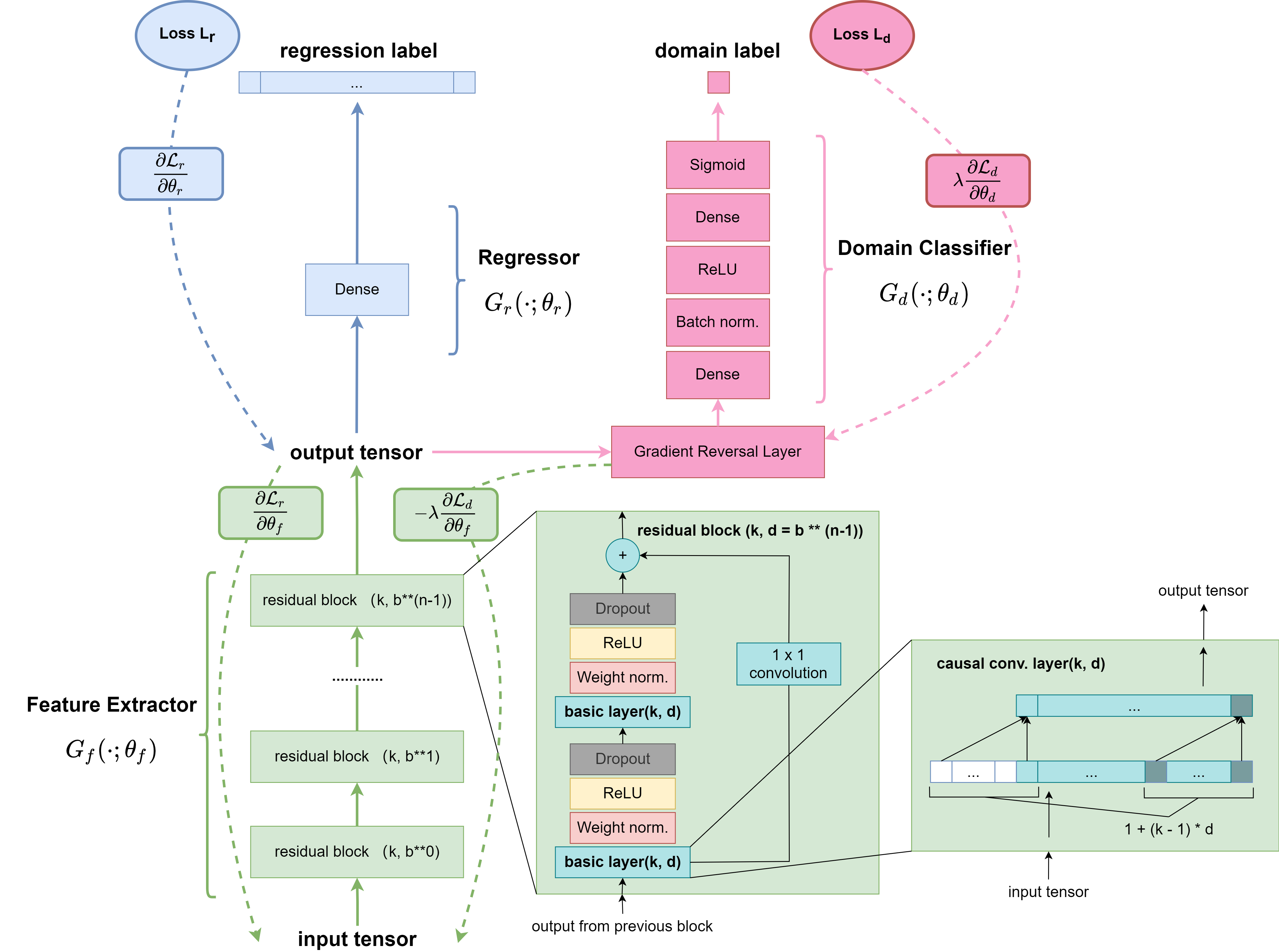}}
\caption{Temporal Convolution Networks Domain Adversarial Architecture}
\label{fig:TCN DANN architecture}
\end{figure*}

In this paper, we modify the DANN framework to make it applicable to our time-series prediction problem. Specifically, a series $[\bm{x}_1,...,\bm{x}_T]$  where each $\bm{x}_i$ is a multi-dimensional vector is input data and a series $[\bm{y}_1,...,\bm{y}_T]$ where each $\bm{y}_i$ is real value is the corresponding output label. The whole architecture we construct is shown in Fig.\ref{fig:TCN DANN architecture}, and it mainly consists of three parts which we shall introduce in detail. 

As shown on the bottom of Fig.\ref{fig:TCN DANN architecture}, the first part is feature extractor denoted by $G_f(\cdot;\theta_f)$ with parameters $\theta_f$. We use Temporal Convolution Networks (TCN) as the backbone of the domain adaptation architecture. With causal and dilated convolutions, TCN can map an input sequence of any length into an output sequence of the same length and greatly increase the reception field within fewer convolutional layers. As shown in Fig.\ref{fig:Dilated Convolution}, the convolution will have some space between the convoluted nodes instead of covering consecutive nodes. With designed dilation size, dilated convolutions help to enlarge the reception field exponentially. This makes TCN show appealing performance with fewer layers. Together with residual connections in the architecture, TCN also improves gradients update when layers of the architecture increase. In our architecture, we use four TCN Residual Block with dilation size increasing exponentially. In each Residual Block, two layers of dilated causal convolution followed by auxiliary layers are stacked together. The output of the last layer will be added to the result of up-sampled input to create a residual connection within the block.

Follow the output of feature extractor and forward to the top of Fig.\ref{fig:TCN DANN architecture}, we will see the second part. To accomplish the goal of prediction, features extracted by TCN will be forward to our regressor denoted by $G_r(\cdot;\theta_r)$  with parameters $\theta_r$. According to \cite{SALINAS20201181}, regressor in our architecture predicts a Gaussian distribution instead of the exact value of the label. The Gaussian likelihood performs better than the mere predicted value when we consider the noise. Consequently, this change will improve the performance of our model. Formly speaking, we clarify that

$$(\mu_i,\sigma_i) = G_r(G_f(\bm{x}_i;\theta_f);\theta_r)$$
and further more, we get regression loss from the prediction likelihood 
$$\mathcal{L}_r((\mu_i,\sigma_i),y_i) = - \frac{1}{\sqrt{2\pi}\sigma_i} \exp{(-\frac{(y_i-\mu_i)^2}{2\sigma_i^2})}$$
where $\mu_i$ and $\sigma_i$ represent mean and standard deviation of predicted Gaussian distribution. $G_f(\cdot;\theta_f)$ represents the feature extractor with parameters $\theta_f$. $G_r(\cdot;\theta_r)$ represents the regressor with parameters $\theta_r$.

To the right of regressor in Fig.\ref{fig:TCN DANN architecture} is a critical part for domain adversary in our architecture. It is a domain classifier denoted by $G_d(\cdot;\theta_d)$  with parameters $\theta_d$. Consisting of several dense layers with batch normalization and ReLU, the domain classifier tends to discriminate features from TCN between source and target domains. On the other side, the feature extractor aims to confuse the discriminator by producing features without domain characteristics. As a result, domain adversarial training between feature extractor and domain classifier helps the model adapt to the target domain and increase prediction accuracy.

In general, we will get the prediction loss and the domain loss respectively by

$$\mathcal{L}_r^i(\theta_f,\theta_r) = \mathcal{L}_r(G_r(G_f(\bm{x}_i;\theta_f);\theta_r),y_i)$$

$$\mathcal{L}_d^i(\theta_f,\theta_d) = \mathcal{L}_d(G_d(G_f(\bm{x}_i;\theta_f);\theta_d),d_i)$$

Where $d_i \in \{0,1\}$ represent the domain label of the input data $\bm{x}_i$, additionally, $\mathcal{L}_r^i$ represents the loss computed from the regressor as mentioned above. $\mathcal{L}_d^i$ represents the loss computed from the domain classifier with the Binary Cross-Entropy loss function.

To train the DANN networks, we aim to optimize

$$E(\theta_f,\theta_r,\theta_d) = \frac{1}{T}\sum_{i=1}^{T}(\mathcal{L}_r^{i}(\theta_f,\theta_r) - \lambda\mathcal{L}_d^i(\theta_f,\theta_d))$$
by finding the saddle point $\hat{\theta}_f$,$\hat{\theta}_r$,$\hat{\theta}_d$ such that
\begin{align*}
    (\hat{\theta}_f,\hat{\theta}_r) \quad &=\quad \mathop{\arg\min}\limits_{\theta_f,\theta_r} E(\theta_f,\theta_r,\hat{\theta_d}) \\
    \hat{\theta}_d \quad&=\quad \mathop{\arg\max}\limits_{\theta_d} E(\hat{\theta_f},\hat{\theta_r},\theta_d)
\end{align*}

In the process of training, gradient updates are employed to find the target point as follows:
\begin{align*}
    \theta_f \quad &\longleftarrow \quad \theta_f - \mu (\frac{\partial\mathcal{L}_r^i}{\partial\theta_f} - \lambda \frac{\partial \mathcal{L}_d^i}{\partial\theta_f}) \\ 
    \theta_r \quad &\longleftarrow \quad \theta_r - \mu  \frac{\partial \mathcal{L}_r^i}{\partial\theta_r}\\ 
    \theta_d \quad &\longleftarrow \quad \theta_d - \mu \lambda \frac{\partial \mathcal{L}_d^i}{\partial\theta_d}
\end{align*}
where $\mu$ is the learning rate which was set to $10^{-3}$ in our experiments.

We note that the gradients from regression loss and domain loss are subtracted, which is different from most cases in modern deep learning framework. With a gradient reversal layer (GRL) \cite{ganin2016domain}, the standard optimization algorithm in the framework can be utilized in training. As shown in Fig.\ref{fig:TCN DANN architecture}, domain loss $\mathcal{L}_d$ will backward to the domain classifier as usual, but its sign will be reversed at GRL and backward to feature extractor. Consequently, the optimizer in the deep learning framework can be utilized to proceed this gradient updates process. This extra layer makes the implementation of the special domain adversary architecture convenient and straightforward.

In the final stage, the feature extractor will generate domain-invariant features that assist the regressor in accurately predicting the label of input data.
The domain classifier will be removed from the model since the domain label will not be used in the inference period.

\section{EXPERIMENTS}\label{sec:experiment}
\subsection{Experimental Settings}

We conduct all of our experiments on a Linux Server (GPU: NVIDIA RTX 2080). As stated in Section \ref{sec:data}, we generate a large amount of crops data with WOFOST\cite{van1989wofost} as our source domain dataset. Real-world crops data are collected on a farm located in North  China as the target domain dataset. Simulated data covers over 400 complete rounds of maize corn, and each round consists of 173 days. The size of collected real-world data is relatively modest, and it consists of 46 different batches of growth data expanding to 11 days. The considerable gap between source and target domains reflects the difficulty of transfer learning.

\begin{table}[htbp]
\begin{center}
\caption{Parameters Setting of our DANN}
\label{table:1}
\begin{tabular}{c|cc}
\hline
                      & Layer                                                                              & Parameters                                                                                                                  \\ \hline
TCN Feature Extractor & TCN                                                                                & \begin{tabular}[c]{@{}c@{}}input: 5 channels\\ hidden: 40 channels\\ output: 40 channels\\ sequence length: 11\end{tabular} \\ \hline
Regressor             & \begin{tabular}[c]{@{}c@{}}Dense\\ ReLU\end{tabular}                               & \begin{tabular}[c]{@{}c@{}}(40,2)\\\quad \end{tabular}                                                                      \\ \hline
Discriminator         & \begin{tabular}[c]{@{}c@{}}Dense\\ BatchNorm\\ ReLU\\ Dense\\ Sigmoid\end{tabular} & \begin{tabular}[c]{@{}c@{}}(440,64)\\ \\ \\ (64,1)\\ \quad\end{tabular}                                                  \\ \hline
\end{tabular}
\end{center}
\end{table}

\begin{enumerate}
    \item \emph{Evaluation Metric}: Mean Absolute Errors (MAE) and Root Mean Squared Errors (RMSE) are used to evaluate the performance of various models. Though our regressor uses subtracted likelihood as the loss function, we use RMSE to evaluate our approach's performance. The calculation of MAE and RMSE are shown below:
        \begin{align*}
            MAE &=  \frac{1}{TN} \sum_{t=1}^{T}  \sum_{i=1}^{N} \lvert (y_t^i - \widehat{y_t^i}) \rvert \\ 
            RMSE &= \sqrt{ \frac{1}{TN} \sum_{t=1}^{T}  \sum_{i=1}^{N}  (y_t^i - \widehat{y_t^i})^2 }
        \end{align*}
    where $y_t^i$ represents the label for data with index $i$ at timestamp $t$ and $\widehat{y_t^i}$ represents the corresponding predicted label.
    \item \emph{Baseline Models}: For DANN, we implement two variants:
    \begin{itemize}
        \item DANN(LSTM): using LSTM as the feature extractor to learn the mapping from the space of input data to the latent space.
        \item DANN(TCN): using TCN as the feature extractor to learn the mapping from the input data space to the latent space. The structure of TCN is less complicated, but its carefully designed architecture helps extract better latent features.
    \end{itemize}
    We compare DANN with the following baselines:
    \begin{itemize}
        \item MLP (Multilayer Perceptron) (with fine-tuning): a feed-forward neural network that has been widely used in classification and regression problems. Its ability in function approximation has made it effective in neural networks architecture. However, it also relies on feature extraction and the availability of a large amount of data. We implement a MLP with three layers with 40 features in hidden layers. 
        \item LSTM (Long Short-Term Memory) (with fine-tuning): a classic recurrent-based model with a deliberately designed cell unit. It is popular in many time-series predictions. We implement a LSTM with 1, 2, and 4 layers with 40 features in hidden layers.
        \item TCN (Temporal Convolutional Networks) (with fine-tuning): an innovative convolution network architecture with casual and dilated convolutions. We implemented a TCN with 1, 2, and 4 layers with 40 features in hidden layers.
        \item ADDA-LSTM (Adversarial Discriminative Domain Adaption): a domain adaptation architecture that also employs a domain classifier to train a better extractor on the target domain. We use LSTM as the feature extractor on both domains. We implement LSTM with 2 layers with 40 features in hidden layers.
    \end{itemize}
\end{enumerate}

We conduct three sets of experiments based on the baselines mentioned above. In the first experiment (\emph{Feature Extractors}), we train and test existing feature extractors directly on the target domain dataset. This experiment is intended to show the performance of existing feature extractors. For the transfer learning part (\emph{Transfer Learning}), we compare traditional transfer learning and DANN. 
In fine-tuning, we pretrain MLP, LSTM, and TCN on the source domain dataset and retrain them on little data from the target domain. 
Furthermore, we apply our TCN-DANN architecture to the transfer learning problem. 
Finally, we compare DANN's performance 
with LSTM or TCN as the feature extractor (\emph{Ablation Study}).
The parameters setting for our proposed DANN architecture is described in TABLE \ref{table:1}.

\subsection{Experimental Results}
\subsubsection{Feature Extractors}

\begin{table}[htbp]
    \begin{center}
        \caption{Perfomance of Feature Extractor}
        \label{table:feature extractor}
        \setlength{\tabcolsep}{7mm}
        \renewcommand\arraystretch{1.25}
        \begin{tabular}{l|cc}
        \hline
        Model           & \multicolumn{1}{c}{MAE}    & \multicolumn{1}{c}{RMSE}   \\ \hline
        MLP             & 0.006834                   & 0.008003                   \\ 
        LSTM            & 0.001875 &  0.002762 \\ 
        TCN             & 0.002384 & 0.003378 \\ \hline
        \end{tabular}
    \end{center}
\end{table}

In the first experiment, we train some existing models directly on the target domain data. We select two classic approaches (MLP and LSTM) and an innovative network (TCN). 
MLP has the powerful ability to approximate functions.  
Though it does not take the temporal dependency of input data into consideration, it can still accomplish some prediction tasks where the distribution of the dataset is not very complicated. LSTM leverages temporal dependency to extract better features from input data. With deliberately designed unit cells to manipulate the hidden status of the model, LSTM outperforms traditional Recurrent Neural Networks when the length of input data is relatively long. TCN leverages casual convolutions to process time-series data and map the input sequence to a same-length sequence in the latent space. Dilated convolutions and residual connections play a critical role in the architecture of TCN  which show promising performance and potential. Table \ref{table:feature extractor} validates our assumptions, showing that TCN with RMSE loss of 0.003378 and LSTM with RMSE loss of 0.002762. We note that the gap between LSTM and TCN is much smaller than in MLP, which means for time series data, convolution-based TCN is also an appealing method in prediction tasks as against the common default practice of using LSTM or similar architecture. What is more, temporal dependency is proved to be critical in the prediction of time series data since both LSTM and TCN leverage temporal relationships concealed in the dataset distribution, which further confirms our design decision to use TCN, and in the experiments followed, we can see the advantages of using TCN over LSTM.

\subsubsection{Transfer Learning}
\begin{table}[htbp]
    \begin{center}
        \caption{Perfomance of Transfer Learning}
        \label{table:transfer learning}
        \setlength{\tabcolsep}{7mm}
        \renewcommand\arraystretch{1.1}
        \begin{tabular}{l|ll}
        \hline
        Model           & \multicolumn{1}{c}{MAE}    & \multicolumn{1}{c}{RMSE}   \\ \hline
        MLP-Finetuning  & 0.009752                   & 0.011633                   \\ 
        LSTM-Finetuning & \textbf{0.001689} & 0.002676\\
        TCN-Finetuning  & 0.001781 & \textbf{0.002621}\\ \hline
        DANN-LSTM       & \textbf{0.001925} & 0.002976\\ 
        DANN-TCN        & 0.001964 & \textbf{0.002789}\\ \hline
        ADDA-LSTM  & 0.016917                   & 0.020881                  \\ \hline
        \end{tabular}
    \end{center}
\end{table}

\begin{table*}[htbp]
    \begin{center}
        \caption{Perfomance Comparison Between LSTM and TCN}
        \label{table:LSTM and TCN}
        \setlength{\tabcolsep}{7mm}
        \renewcommand\arraystretch{1.1}
        \begin{tabular}{l|ll}
        \hline
        Model           & \multicolumn{1}{c}{MAE (1 / 2 / 4 Layers)}    & \multicolumn{1}{c}{RMSE (1 / 2 / 4 Layers)}   \\ \hline
        LSTM            & 0.003374\ /\ 0.002031\ /\ 0.001875 & 0.004424\ /\ 0.002994\ /\ 0.002762 \\ 
        LSTM-Finetuning & 0.001689\ /\ 0,001801\ /\ 0.001855 & 0.002676\ /\ 0.002686\ /\ 0.002740 \\
        DANN-LSTM       & 0.001944\ /\ 0.001925\ /\ 0.002708 & 0.003007\ /\ 0.002976\ /\ 0.003791 \\ \hline
        TCN             & 0.002924\ /\ 0.003826\ /\ 0.003048 & 0.003772\ /\ 0.004719\ /\ 0.004079 \\ 
        TCN-Finetuning  & 0.001929\ /\ \textbf{0.001781}\ /\ 0.001829 & 0.002799\ /\ \textbf{0.002621}\ /\ 0.002652 \\ 
        DANN-TCN        & 0.001943\ /\ 0.001964\ /\ 0.001956 & 0.002821\ /\ \textbf{0.002789}\ /\ 0.002885 \\ \hline
        
        \end{tabular}
    \end{center}
\end{table*}

To verify the effectiveness of our proposed domain adaptation architecture, then we conduct fine-tuning of these approaches and two variants of DANN on our crop dataset. Different from traditional deep learning, where outstanding performances depend on a large amount of data, few training samples are available in our target domain. In the fine-tuning process, models are first trained on source domain data. Later, parameters are transferred to the target domain and retrained to improve the accuracy of the model. While in our proposed domain adaptation architecture, domain adversarial training helped to accomplish the same goal of fine-tuning. Table \ref{table:transfer learning} reveals that transfer learning is effective since the information from the source domain is exploited to increase model performance in the target domain. As shown in Table \ref{table:transfer learning}, we note that TCN backboned architecture outperforms other models with RMSE loss up to 0.002621 and 0.002789. This result validates that DANN architecture is able to give out domain-invariant features and is good at transfer learning since the gap between DANN and fine-tuning is so narrow. What is more, we also construct an Adversarial Discriminative Domain Adaption (ADDA) architecture using LSTM to compare different domain adaptation architectures. We believe that ADDA is also a powerful model of transfer learning, but it is more complicated in the implementation and search of hyperparameters. Our LSTM based ADDA did not perform as expected. However, DANN is also a complicated architecture compared to a mere fine-tuning. 
We observe only a narrow gap between the performance of DANN and fine-tuning models (specifically for two TCN based networks). We believe that the dimensionality of the target domain dataset (5 features) is not big enough to exploit the full potential of DANN based architecture.  But our results show that even with such limited features, our heavy-weight DANN has a very close performance of fine-tuning (still better than MLP finetuning) and much better than an equally heavy-weight domain adaptation network (ADDA-LSTM).
We hypothesize that when applied in the real world where the underlying data is more complex, our proposed architecture shall outperform the method of fine-tuning. We will conduct this as a future work for other pervasive application systems with similar issues with high-quality data shortages.

\subsubsection{
Ablation Study of TCN and LSTM}
\begin{figure}[htbp]

\centerline{\includegraphics[width=0.45\textwidth]{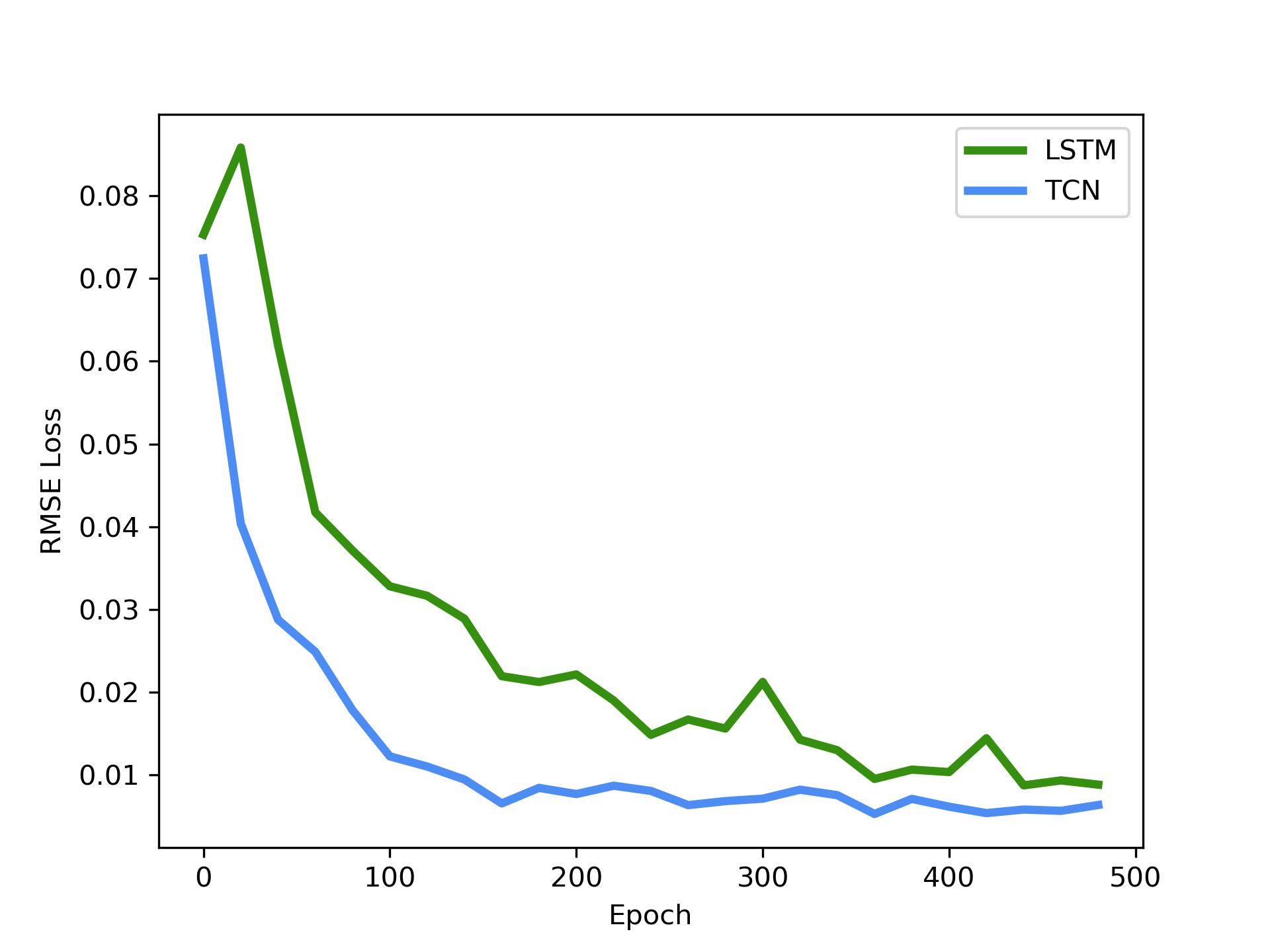}}
\caption{RMSE Loss of DANN-TCN and DANN-LSTM (4 layers)}
\label{fig:RMSE Loss}
\end{figure}

\begin{table}[htbp]
\begin{center}
\caption{Parameters Size}
\label{table:parameterts size}
\setlength{\tabcolsep}{7mm}
\renewcommand\arraystretch{1.5}
\begin{tabular}{l|l}
\hline
Model & Parameters(1 / 2 / 4 Layers) \\ \hline
LSTM  & 7,520\ /\ 20,640\ /\ 46,880      \\
TCN   & 4,000\ /\ 10,560\ /\ 23,680      \\ \hline
\end{tabular}
\end{center}
\end{table}

In our last setting, we compare TCN and LSTM in the same domain adaptation architecture. We construct DANN with TCN and LSTM ,respectively in 1, 2, and 4 layers. From Table \ref{table:LSTM and TCN}, we notice that TCN-based outperforms LSTM-based in the case of accuracy in time series data predictions with the lowest RMSE of 0.002621. This validates our design choice of using TCN in our domain adaptation network and shows that TCN is a promising architecture in time series prediction tasks in general. Additionally, TCN bears fewer parameters when compared to LSTM with the same layers as shown in Table \ref{table:parameterts size}. TCN requires almost half of the size of LSTM network architecture, which makes TCN a much suitable candidate for pervasive computing applications where computation and storage are constraints. Moreover, the gap in the model size most likely will result in different convergence rates. From Fig.\ref{fig:RMSE Loss}, we note that TCN based DANN converged after nearly 150 epochs of training while it took LSTM based DANN over 500 epochs to stabilize its RMSE loss.  This is another insightful finding as for those pervasive systems which require continual learning (autonomous driving models), the convergence rate is crucial. 

In summary, we conduct three sets of experiments with simulated crops data generated from WOFOST and real-world crops data collected from a real-world farm. The experimental results show that our proposed TCN backboned DANN architecture outperforms baselines in general. Moreover, we notice the promising future for TCN to be extensively studied and used for pervasive application systems for its better accuracy, smaller model size, and faster convergence rate.

\section{conclusion and discussion}\label{sec:conclusion}

In this paper, we have proposed a TCN backboned DANN transfer learning architecture to predict crops growth with simulated data as the source domain dataset and collected real-world data as the target domain dataset. We have conducted a comprehensive analysis of our proposed model and compared it with the state-of-the-art transfer learning approach.
The experiment shows promising results of our proposed architecture compared with other benchmarks. Specifically, we find that TCN has better accuracy, a smaller model size, and a faster convergence rate. As future work, we will apply our TCN based domain adaptation framework for diverse pervasive applications with similar training data constraints.


\newpage

\bibliographystyle{IEEEtran}
\bibliography{references}

\begin{thebibliography}{10}
\providecommand{\url}[1]{#1}
\csname url@samestyle\endcsname
\providecommand{\newblock}{\relax}
\providecommand{\bibinfo}[2]{#2}
\providecommand{\BIBentrySTDinterwordspacing}{\spaceskip=0pt\relax}
\providecommand{\BIBentryALTinterwordstretchfactor}{4}
\providecommand{\BIBentryALTinterwordspacing}{\spaceskip=\fontdimen2\font plus
\BIBentryALTinterwordstretchfactor\fontdimen3\font minus
  \fontdimen4\font\relax}
\providecommand{\BIBforeignlanguage}[2]{{%
\expandafter\ifx\csname l@#1\endcsname\relax
\typeout{** WARNING: IEEEtran.bst: No hyphenation pattern has been}%
\typeout{** loaded for the language `#1'. Using the pattern for}%
\typeout{** the default language instead.}%
\else
\language=\csname l@#1\endcsname
\fi
#2}}
\providecommand{\BIBdecl}{\relax}
\BIBdecl

\bibitem{horie1992yield}
T.~Horie, M.~Yajima, and H.~Nakagawa, ``Yield forecasting,'' \emph{Agricultural
  systems}, vol.~40, no. 1-3, pp. 211--236, 1992.

\bibitem{li2017oryza2000}
T.~Li, O.~Angeles, M.~Marcaida~III, E.~Manalo, M.~P. Manalili, A.~Radanielson,
  and S.~Mohanty, ``From oryza2000 to oryza (v3): An improved simulation model
  for rice in drought and nitrogen-deficient environments,'' \emph{Agricultural
  and forest meteorology}, vol. 237, pp. 246--256, 2017.

\bibitem{van1989wofost}
C.~v. Van~Diepen, J.~Wolf, H.~Van~Keulen, and C.~Rappoldt, ``Wofost: a
  simulation model of crop production,'' \emph{Soil use and management},
  vol.~5, no.~1, pp. 16--24, 1989.

\bibitem{jones2003dssat}
J.~W. Jones, G.~Hoogenboom, C.~H. Porter, K.~J. Boote, W.~D. Batchelor,
  L.~Hunt, P.~W. Wilkens, U.~Singh, A.~J. Gijsman, and J.~T. Ritchie, ``The
  dssat cropping system model,'' \emph{European journal of agronomy}, vol.~18,
  no. 3-4, pp. 235--265, 2003.

\bibitem{WHISLER1986141}
F.~Whisler, B.~Acock, D.~Baker, R.~Fye, H.~Hodges, J.~Lambert, H.~Lemmon,
  J.~McKinion, and V.~Reddy, ``Crop simulation models in agronomic systems,''
  ser. Advances in Agronomy, N.~Brady, Ed.\hskip 1em plus 0.5em minus
  0.4em\relax Academic Press, 1986, vol.~40, pp. 141--208.

\bibitem{wang2020deep}
H.~Wang, E.~Cimen, N.~Singh, and E.~Buckler, ``Deep learning for plant genomics
  and crop improvement,'' \emph{Current opinion in plant biology}, vol.~54, pp.
  34--41, 2020.

\bibitem{10.1145/3209811.3212707}
\BIBentryALTinterwordspacing
A.~X. Wang, C.~Tran, N.~Desai, D.~Lobell, and S.~Ermon, ``Deep transfer
  learning for crop yield prediction with remote sensing data,'' in
  \emph{Proceedings of the 1st ACM SIGCAS Conference on Computing and
  Sustainable Societies}, ser. COMPASS '18.\hskip 1em plus 0.5em minus
  0.4em\relax New York, NY, USA: Association for Computing Machinery, 2018.
  [Online]. Available: \url{https://doi.org/10.1145/3209811.3212707}
\BIBentrySTDinterwordspacing

\bibitem{VANKLOMPENBURG2020105709}
\BIBentryALTinterwordspacing
T.~{van Klompenburg}, A.~Kassahun, and C.~Catal, ``Crop yield prediction using
  machine learning: A systematic literature review,'' \emph{Computers and
  Electronics in Agriculture}, vol. 177, p. 105709, 2020. [Online]. Available:
  \url{https://www.sciencedirect.com/science/article/pii/S0168169920302301}
\BIBentrySTDinterwordspacing

\bibitem{keating2003overview}
B.~A. Keating, P.~S. Carberry, G.~L. Hammer, M.~E. Probert, M.~J. Robertson,
  D.~Holzworth, N.~I. Huth, J.~N. Hargreaves, H.~Meinke, Z.~Hochman
  \emph{et~al.}, ``An overview of apsim, a model designed for farming systems
  simulation,'' \emph{European journal of agronomy}, vol.~18, no. 3-4, pp.
  267--288, 2003.

\bibitem{cambra2019smart}
C.~Cambra~Baseca, S.~Sendra, J.~Lloret, and J.~Tomas, ``A smart decision system
  for digital farming,'' \emph{Agronomy}, vol.~9, no.~5, p. 216, 2019.

\bibitem{salam2019internet}
A.~Salam and S.~Shah, ``Internet of things in smart agriculture: Enabling
  technologies,'' in \emph{2019 IEEE 5th World Forum on Internet of Things
  (WF-IoT)}.\hskip 1em plus 0.5em minus 0.4em\relax IEEE, 2019, pp. 692--695.

\bibitem{sun2019county}
J.~Sun, L.~Di, Z.~Sun, Y.~Shen, and Z.~Lai, ``County-level soybean yield
  prediction using deep cnn-lstm model,'' \emph{Sensors}, vol.~19, no.~20, p.
  4363, 2019.

\bibitem{khaki2020cnn}
S.~Khaki, L.~Wang, and S.~V. Archontoulis, ``A cnn-rnn framework for crop yield
  prediction,'' \emph{Frontiers in Plant Science}, vol.~10, p. 1750, 2020.

\bibitem{wang2020winter}
X.~Wang, J.~Huang, Q.~Feng, and D.~Yin, ``Winter wheat yield prediction at
  county level and uncertainty analysis in main wheat-producing regions of
  china with deep learning approaches,'' \emph{Remote Sensing}, vol.~12,
  no.~11, p. 1744, 2020.

\bibitem{drummond2003statistical}
S.~T. Drummond, K.~A. Sudduth, A.~Joshi, S.~J. Birrell, and N.~R. Kitchen,
  ``Statistical and neural methods for site--specific yield prediction,''
  \emph{Transactions of the ASAE}, vol.~46, no.~1, p.~5, 2003.

\bibitem{fortin2011site}
J.~G. Fortin, F.~Anctil, L.-{\'E}. Parent, and M.~A. Bolinder, ``Site-specific
  early season potato yield forecast by neural network in eastern canada,''
  \emph{Precision Agriculture}, vol.~12, no.~6, pp. 905--923, 2011.

\bibitem{russ2009data}
G.~Ru{\ss}, ``Data mining of agricultural yield data: A comparison of
  regression models,'' in \emph{Industrial Conference on Data Mining}.\hskip
  1em plus 0.5em minus 0.4em\relax Springer, 2009, pp. 24--37.

\bibitem{zhang2010simulation}
L.~Zhang, J.~Zhang, S.~Kyei-Boahen, M.~Zhang \emph{et~al.}, ``Simulation and
  prediction of soybean growth and development under field conditions,''
  \emph{Am-Euras J Agr Environ Sci}, vol.~7, no.~4, pp. 374--385, 2010.

\bibitem{gonzalez2014predictive}
A.~Gonz{\'a}lez~S{\'a}nchez, J.~Frausto~Sol{\'\i}s, W.~Ojeda~Bustamante
  \emph{et~al.}, ``Predictive ability of machine learning methods for massive
  crop yield prediction,'' 2014.

\bibitem{niedbala2019application}
G.~Niedba{\l}a, ``Application of artificial neural networks for multi-criteria
  yield prediction of winter rapeseed,'' \emph{Sustainability}, vol.~11, no.~2,
  p. 533, 2019.

\bibitem{schwalbert2020satellite}
R.~A. Schwalbert, T.~Amado, G.~Corassa, L.~P. Pott, P.~V. Prasad, and I.~A.
  Ciampitti, ``Satellite-based soybean yield forecast: Integrating machine
  learning and weather data for improving crop yield prediction in southern
  brazil,'' \emph{Agricultural and Forest Meteorology}, vol. 284, p. 107886,
  2020.

\bibitem{jiang2020deep}
H.~Jiang, H.~Hu, R.~Zhong, J.~Xu, J.~Xu, J.~Huang, S.~Wang, Y.~Ying, and
  T.~Lin, ``A deep learning approach to conflating heterogeneous geospatial
  data for corn yield estimation: A case study of the us corn belt at the
  county level,'' \emph{Global change biology}, vol.~26, no.~3, pp. 1754--1766,
  2020.

\bibitem{zhang2020combining}
L.~Zhang, Z.~Zhang, Y.~Luo, J.~Cao, and F.~Tao, ``Combining optical,
  fluorescence, thermal satellite, and environmental data to predict
  county-level maize yield in china using machine learning approaches,''
  \emph{Remote Sensing}, vol.~12, no.~1, p.~21, 2020.

\bibitem{zadrozny2004learning}
B.~Zadrozny, ``Learning and evaluating classifiers under sample selection
  bias,'' in \emph{Proceedings of the twenty-first international conference on
  Machine learning}, 2004, p. 114.

\bibitem{kanamori2008efficient}
T.~Kanamori, S.~Hido, and M.~Sugiyama, ``Efficient direct density ratio
  estimation for non-stationarity adaptation and outlier detection,'' in
  \emph{Advances in neural information processing systems}.\hskip 1em plus
  0.5em minus 0.4em\relax Citeseer, 2008, pp. 809--816.

\bibitem{yang2007cross}
J.~Yang, R.~Yan, and A.~G. Hauptmann, ``Cross-domain video concept detection
  using adaptive svms,'' in \emph{Proceedings of the 15th ACM international
  conference on Multimedia}, 2007, pp. 188--197.

\bibitem{duan2012exploiting}
L.~Duan, D.~Xu, and S.-F. Chang, ``Exploiting web images for event recognition
  in consumer videos: A multiple source domain adaptation approach,'' in
  \emph{2012 IEEE Conference on computer vision and pattern recognition}.\hskip
  1em plus 0.5em minus 0.4em\relax IEEE, 2012, pp. 1338--1345.

\bibitem{daume2009frustratingly}
H.~Daum{\'e}~III, ``Frustratingly easy domain adaptation,'' \emph{arXiv
  preprint arXiv:0907.1815}, 2009.

\bibitem{fernando2013unsupervised}
B.~Fernando, A.~Habrard, M.~Sebban, and T.~Tuytelaars, ``Unsupervised visual
  domain adaptation using subspace alignment,'' in \emph{Proceedings of the
  IEEE international conference on computer vision}, 2013, pp. 2960--2967.

\bibitem{pan2010domain}
S.~J. Pan, I.~W. Tsang, J.~T. Kwok, and Q.~Yang, ``Domain adaptation via
  transfer component analysis,'' \emph{IEEE transactions on neural networks},
  vol.~22, no.~2, pp. 199--210, 2010.

\bibitem{yosinski2014transferable}
J.~Yosinski, J.~Clune, Y.~Bengio, and H.~Lipson, ``How transferable are
  features in deep neural networks?'' \emph{arXiv preprint arXiv:1411.1792},
  2014.

\bibitem{ganin2016domain}
Y.~Ganin, E.~Ustinova, H.~Ajakan, P.~Germain, H.~Larochelle, F.~Laviolette,
  M.~Marchand, and V.~Lempitsky, ``Domain-adversarial training of neural
  networks,'' \emph{The journal of machine learning research}, vol.~17, no.~1,
  pp. 2096--2030, 2016.

\bibitem{tzeng2017adversarial}
E.~Tzeng, J.~Hoffman, K.~Saenko, and T.~Darrell, ``Adversarial discriminative
  domain adaptation,'' in \emph{Proceedings of the IEEE conference on computer
  vision and pattern recognition}, 2017, pp. 7167--7176.

\bibitem{goodfellow2014generative}
I.~Goodfellow, J.~Pouget-Abadie, M.~Mirza, B.~Xu, D.~Warde-Farley, S.~Ozair,
  A.~Courville, and Y.~Bengio, ``Generative adversarial nets,'' \emph{Advances
  in neural information processing systems}, vol.~27, 2014.

\bibitem{lea2017temporal}
C.~Lea, M.~D. Flynn, R.~Vidal, A.~Reiter, and G.~D. Hager, ``Temporal
  convolutional networks for action segmentation and detection,'' in
  \emph{proceedings of the IEEE Conference on Computer Vision and Pattern
  Recognition}, 2017, pp. 156--165.

\bibitem{bai2018empirical}
S.~Bai, J.~Z. Kolter, and V.~Koltun, ``An empirical evaluation of generic
  convolutional and recurrent networks for sequence modeling,'' \emph{arXiv
  preprint arXiv:1803.01271}, 2018.

\bibitem{AnnualClimateChina19812010}
N.~M. S.~D. Center, ``Annual data set of standard values of surface climate in
  china (1981-2010),''
  \url{http://data.cma.cn/data/cdcdetail/dataCode/SURF_CLI_CHN_MUL_MYER_19812010.html}.

\bibitem{shao2021few-shot}
W.~Shao, S.~Zhao, Z.~Zhang, S.~Wang, M.~S. Rahaman, A.~Song, and F.~D. Salim,
  ``Fadacs: A few-shot adversarial domain adaptation architecture for
  context-aware parking availability sensing,'' in \emph{2021 IEEE
  International Conference on Pervasive Computing and Communications (PerCom)},
  2021, pp. 1--10.

\bibitem{SALINAS20201181}
\BIBentryALTinterwordspacing
D.~Salinas, V.~Flunkert, J.~Gasthaus, and T.~Januschowski, ``Deepar:
  Probabilistic forecasting with autoregressive recurrent networks,''
  \emph{International Journal of Forecasting}, vol.~36, no.~3, pp. 1181--1191,
  2020. [Online]. Available:
  \url{https://www.sciencedirect.com/science/article/pii/S0169207019301888}
\BIBentrySTDinterwordspacing

\end{thebibliography}

\end{document}